\documentclass[conference]{IEEEtran}
\usepackage[T1]{fontenc}
\usepackage{textcomp}
\usepackage{amsmath,amsfonts,amssymb}
\usepackage{graphicx}
\usepackage{booktabs}
\usepackage{hyperref}
\usepackage{siunitx}
\usepackage{pifont}
\usepackage{adjustbox} 
\usepackage{longtable}
\usepackage{tikz}
\usetikzlibrary{shapes.geometric, arrows.meta, positioning, fit, calc, shapes.misc, shadows}
\usepackage{tabularx}
\usepackage{ragged2e} 
\usepackage{supertabular}
\usepackage{array} 
\usepackage{booktabs} 
\usepackage{caption}
\usepackage{subcaption}

\newcolumntype{L}{>{\RaggedRight\arraybackslash}X}

\usepackage{breqn}
\usepackage{tabularx}   

\title{Accelerating HEC-RAS: A Recurrent Neural Operator for Rapid River Forecasting}

\author{
  \IEEEauthorblockN{Edward Holmberg}
  \IEEEauthorblockA{\footnotesize\textit{Gulf States Center for Env. Informatics}\\
                   \textit{University of New Orleans}\\
                   Louisiana, USA\\
                   eholmber@uno.edu}
  \and
  \IEEEauthorblockN{Pujan Pokhrel}
  \IEEEauthorblockA{\footnotesize\textit{Gulf States Center for Env. Informatics}\\
                   \textit{University of New Orleans}\\
                   Louisiana, USA\\
                   ppokhre1@uno.edu}
  \and
  \IEEEauthorblockN{Maximilian Zoch}
  \IEEEauthorblockA{\footnotesize \textit{CoDiS-Lab ISDS} \\
    \textit{Graz Technical University of Technology} \\
    Graz, Austria \\
    maximilian.zoch@tugraz.at}

  \and
  \IEEEauthorblockN{Elias Ioup}
  \IEEEauthorblockA{\footnotesize\textit{Center for Geospatial Sciences}\\
                   \textit{Naval Research Laboratory}\\
                   Mississippi, USA\\
                   elias.z.ioup.civ@us.navy.mil}
  \and
  \IEEEauthorblockN{Ken Pathak}
  \IEEEauthorblockA{\footnotesize\textit{US Army Corps of Engineers}\\
                   \textit{Vicksburg District}\\
                   Mississippi, USA\\
                   ken.pathak@usace.army.mil}
  \and
  \IEEEauthorblockN{Steven Sloan}
  \IEEEauthorblockA{\footnotesize\textit{US Army Corps of Engineers}\\
                   \textit{Vicksburg District}\\
                   Mississippi, USA\\
                   steven.sloan@usace.army.mil}
  \and
  \IEEEauthorblockN{Kendall Niles}
  \IEEEauthorblockA{\footnotesize\textit{US Army Corps of Engineers}\\
                   \textit{Vicksburg District}\\
                   Mississippi, USA\\
                   kendall.niles@usace.army.mil}
  \and
  \IEEEauthorblockN{Jay Ratcliff}
  \IEEEauthorblockA{\footnotesize\textit{Gulf States Center for Env. Informatics}\\
                   \textit{University of New Orleans}\\
                   Louisiana, USA\\
                   jratclif@uno.edu}
  \and
  \IEEEauthorblockN{Maik Flanagin}
  \IEEEauthorblockA{\footnotesize\textit{US Army Corps of Engineers}\\
                   \textit{New Orleans District}\\
                   Louisiana, USA\\
                   maik.c.flanagin@usace.army.mil}
  \and
  \IEEEauthorblockN{Christian Guetl}
  \IEEEauthorblockA{\footnotesize\textit{CoDiS-Lab ISDS}\\
                   \textit{Graz University of Technology}\\
                   Graz, Austria\\
                   c.guetl@tugraz.at}
    \and
    \IEEEauthorblockN{Julian Simeonov}
    \IEEEauthorblockA{\footnotesize\textit{Ocean Sciences Division}\\
                     \textit{Naval Research Laboratory}\\
                     Mississippi, USA\\
                     julian.a.simeonov.civ@us.navy.mil}
      \and
  \IEEEauthorblockN{Mahdi Abdelguerfi}
  \IEEEauthorblockA{\footnotesize\textit{Gulf States Center for Env. Informatics}\\
                   \textit{University of New Orleans}\\
                   Louisiana, USA\\
                   gulfsceidirector@uno.edu}
}

\begin{document}
\maketitle

\begin{abstract}
Physics-based solvers like HEC-RAS provide high-fidelity river forecasts but are too computationally intensive for on-the-fly decision-making during flood events. The central challenge is to accelerate these simulations without sacrificing accuracy. This paper introduces a deep learning surrogate that treats HEC-RAS not as a solver but as a data-generation engine. We propose a hybrid, auto-regressive architecture that combines a Gated Recurrent Unit (GRU) to capture short-term temporal dynamics with a Geometry-Aware Fourier Neural Operator (Geo-FNO) to model long-range spatial dependencies along a river reach. The model learns underlying physics implicitly from a minimal eight-channel feature vector encoding dynamic state, static geometry, and boundary forcings extracted directly from native HEC-RAS files. Trained on 67 reaches of the Mississippi River Basin, the surrogate was evaluated on a year-long, unseen hold-out simulation. Results show the model achieves a strong predictive accuracy, with a median absolute stage error of 0.31 feet. Critically, for a full 67-reach ensemble forecast, our surrogate reduces the required wall-clock time from 139 minutes to 40 minutes, a speedup of nearly 3.5 times over the traditional solver. The success of this data-driven approach demonstrates that robust feature engineering can produce a viable, high-speed replacement for conventional hydraulic models, improving the computational feasibility of large-scale ensemble flood forecasting.
\end{abstract}


\section{Introduction}\label{sec:intro}

\textbf{The Urgency of Flood Forecasting.}
During a flood, the U.S. Army Corps of Engineers (USACE) must make critical decisions, 
from issuing evacuation orders to scheduling gate operations, within minutes%
\cite{hecras_runtime_2022}. %
This operational tempo is fundamentally at odds with the
$\mathcal{O}(\mathrm{hours})$ wall-clock times required by physics-based solvers
such as \textsc{HEC-RAS} to simulate unsteady flow%
\cite{hecras_runtime_2022,chanson2004}. %
While pre-computed scenario libraries or reduced-order models offer one
work-around, they are often too coarse to capture the specific hydrograph that
unfolds in real time%
\cite{takbiri2020,benner2015}.

\vspace{4pt}
\noindent\textbf{The Challenge.}
The central challenge is to deliver the fidelity of a \textsc{HEC-RAS} simulation\,\cite{hecras_runtime_2022}
at a speed that enables rapid, on-the-fly ensemble forecasting, turning hours of computation into timely
insights\,\cite{hecras_runtime_2022,chanson2004}.

\vspace{4pt}
\noindent\textbf{This Paper's Contribution.}
We address this challenge by reframing the HEC-RAS workflow itself.  Instead of relying on its iterative
solver\,\cite{hecras_runtime_2022}, we treat its native project files as a direct source of training data
for a deep-learning surrogate.  We propose an autoregressive \textit{GRU–GeoFNO} model,
that learns the complex spatio-temporal dynamics of river flow.  The network ingests a minimal eight-channel
vector representing dynamic state, static geometry, and boundary forcings, and then rolls the simulation
forward hour-by-hour.

\vspace{4pt}
\noindent\textbf{Key Innovations:}
\begin{enumerate}
  \item \textit{A True Plug-in Surrogate:} Our model requires no re-meshing or data conversion, reading the native \texttt{.g\#\#}, \texttt{.u\#\#}, and DSS file bundle directly.
  \item \textit{A Minimalist Universal Interface:} We identify a compact eight-channel feature set sufficient for stable, multi-day forecasts that is common across public 1-D HEC-RAS projects.
  \item \textit{A Hybrid Autoregressive Architecture:} Our GRU–GeoFNO loop couples a Gated Recurrent Unit for short-term memory with a Fourier Neural Operator for long-range spatial dependencies, maintaining stability without explicit physics-based loss terms.
\end{enumerate}

\noindent Together, these advances elevate autoregressive neural operators from academic prototypes to operationally promising engines for rapid ensemble flood guidance.


\section{Background: HEC-RAS as a Data Preprocessor}\label{sec:background}

\subsection{HEC-RAS: The Industry-Standard Solver}
HEC-RAS, the U.S. Army Corps of Engineers’ River Analysis System, is widely regarded as the industry-standard platform for river hydraulics
\cite{hecras_runtime_2022}.  Under the hood, it solves the one-dimensional Saint-Venant equations\cite{chanson2004} using an implicit Newton–Raphson finite-difference scheme, with several inner iterations per global time step to balance continuity and momentum
\cite{hecras_runtime_2022}.  This strategy delivers high numerical accuracy but at a steep computational cost: full-reach unsteady-flow simulations typically require hours to days of wall-clock time
\cite{hecras_runtime_2022,takbiri2020}.

\subsection{Novel Use Case: From Solver to Pre-Processor}
This work treats HEC-RAS not as an end-to-end simulation tool, but as a powerful data-generation engine.  By leveraging its mature GIS and project-management capabilities\,\cite{hecras_runtime_2022}, we can assemble consistent geometries, meshes, and boundary hydrographs directly from the native project bundle\,\cite{hecras_runtime_2022,flanagin2007,wilson2003gidb, abdelguerfi2001ser}.  We
then export this curated file set into a machine-learning pipeline.  The surrogate ingests these inputs, learns the hydraulic relationships and returns reach-scale forecasts in seconds rather than hours\,\cite{takbiri2020}.  In this workflow HEC-RAS becomes a build tool for high-quality training data, while the surrogate supplies the speed needed for rapid what-if analyses.

\subsection{The HEC-RAS File Ecosystem}
The key to this approach is the structured, information-rich file bundle that constitutes a standard HEC-RAS project. These files contain all the static, quasi-static, and dynamic information required to train a robust surrogate model, as summarized in Table~\ref{tab:ras_files}.

\begin{table}[htbp]
  \footnotesize
  \centering
  \caption{\textsc{HEC\mbox{-}RAS} file bundle organised by information type. `\#\#` denotes version indices.}

  \label{tab:ras_files}
  \begin{tabularx}{\columnwidth}{@{} l l L @{}} 
    \toprule
    \textbf{Files} & \textbf{Class} & \textbf{Key contents} \\
    \midrule
    
    \multicolumn{3}{@{}l}{\textit{Static geometry}} \\
    \cmidrule(r){1-3} 
    \texttt{*.g\#\#} & XS/1‑D & Station–elevation pairs, banks, centre‑line \\
    \texttt{*.c\#\#} & 2‑D mesh & Cell polygons, bed elevation, roughness zones \\
    \texttt{*.b\#\#} & 1‑D structs & Bridge and culvert shapes, pier spacing \\
    \addlinespace
    
    \multicolumn{3}{@{}l}{\textit{Quasi‑static metadata}} \\
    \cmidrule(r){1-3}
    \texttt{*.p\#\#} & Plan & Geometry/flow linkage, solver tolerances \\
    \addlinespace
    
    \multicolumn{3}{@{}l}{\textit{Dynamic time‑series}} \\
    \cmidrule(r){1-3}
    \texttt{*.u\#\#} & Unsteady flow & Hydrograph pointers, gate schedules, run window \\
    \texttt{*.dss} & DSS & Upstream $Q(t)$, downstream $H(t)$, lateral inflows \\
    
    \bottomrule
  \end{tabularx}
\end{table}

\subsection{Key Hydraulic Terminology}
To interpret the model inputs and outputs, we define the following core terms:
\begin{description}
  \item[Reach]\hfill\\
        A contiguous channel segment between two network break-points
        (e.g.\ a confluence or control structure)\cite{hecras_runtime_2022}.
        Our model operates on a single reach at a time, advancing from an
        \emph{upstream node} (boundary inflow $Q_{\text{up}}$) to a
        \emph{downstream node} (boundary stage $H_{\text{dn}}$).

  \item[Stage ($H$)]\hfill\\
        The water-surface elevation at a cross-section, referenced to a project
        datum such as NAVD 88\cite{chanson2004}.  Units: metres.

  \item[Discharge ($Q$)]\hfill\\
        The volumetric flow rate through a cross-section, defined as positive
        in the downstream direction\cite{chanson2004}.  Units:
        m$^{3}\,\text{s}^{-1}$.
\end{description}

\section{Related Work}\label{sec:related_work}

Our work builds on advances in three key areas: data-driven hydraulic modeling, autoregressive sequence prediction, and neural operators for scientific computing.

\subsection{Data-Driven Surrogates for River Hydraulics}
Early data-driven surrogates for river hydraulics often relied on
feed-forward neural networks or polynomial meta-models to emulate one or two
cross-sections at a time\,\cite{takbiri2020}.
More recent studies have scaled to full reaches by coupling convolutional
encoders with graph neural networks\,\cite{bi2023}, and physics-informed
neural networks have now been demonstrated for single-reach stage
prediction\,\cite{zoch2025pinn}; yet many approaches
remain restricted to steady-flow conditions or simplified rectangular
channels\,\cite{takbiri2020,benner2015}.
In contrast, our study targets the entire unsteady-flow regime of the
Mississippi River model, encompassing 67 distinct reaches and thousands of
irregularly spaced, natural-geometry cross-sections.

\subsection{Autoregressive Models for Temporal Dynamics}
Autoregressive (AR) models, which forecast the next state by feeding back their
own previous outputs, form the backbone of classical time-series
analysis\,\cite{boxjenkins1970}.  The closed-loop structure is computationally
efficient for long-horizon roll-outs, but a known weakness is
\emph{error accumulation}: small mistakes are recycled and amplified,
ultimately drifting the forecast away from reality\,\cite{bengio2015scheduled}.

To mitigate this, modern hydrology has shifted from classical ARMA models to
Recurrent Neural Networks (RNNs) such as Long Short-Term Memory (LSTM) and
Gated Recurrent Units (GRUs)\,\cite{cho2014gru,kratzert2019lstm}.
GRUs use update and reset gates to regulate information flow, capturing
temporal dependencies while remaining parameter-efficient.  When applied
to river networks these RNNs typically predict each gage independently, failing
to capture the spatial physics that connect them\,\cite{kratzert2019lstm}.
Our work addresses this by embedding a GRU within a spatial operator,
allowing the recurrence to span both time and space.  Furthermore, we anchor
the AR loop at every step with the true boundary hydrographs
($Q_{\text{up}},\,H_{\text{dn}}$), providing a strong physical constraint that
drastically reduces long-term drift.

\subsection{Neural Operators for Spatial Dependencies}
To model the spatial physics, we turn to the Fourier Neural Operator (FNO),
which learns mappings between function spaces via global convolutions in the
spectral domain\,\cite{li2021fourier}.  By modulating Fourier modes directly, FNOs capture long-range spatial dependencies with high efficiency and are essentially discretisation invariant\,\cite{li2021fourier, rahaman2019spectral}.  The Geo-FNO
variant extends this concept to irregular meshes by injecting coordinate
information into the spectral block, making it well suited to the
non-uniform cross-section spacing found in river models\,\cite{li2022gno}.
Previous studies have already employed two-dimensional FNOs for rapid
flood-inundation mapping\,\cite{stenta2023inundation}; here we adopt a
one-dimensional Geo-FNO specifically tailored to the chain-like topology of a
river reach.

\subsection{Positioning This Work}
Combining recurrent networks with neural operators is an emerging and powerful tool for modeling complex spatio-temporal systems \cite{brandstetter2022message}. 

A key aspect of our work is its training methodology. We show that the network learns the underlying hydraulic behavior implicitly from the data itself. This is achieved through a carefully engineered eight-channel feature vector that encodes the system's essential physical drivers: the channel geometry ($z_{\text{bed}}, z_{\text{bank}}$), frictional properties ($n_{\text{man}}$), and the mass and energy constraints imposed by boundary hydrographs ($Q_{\text{up}}, H_{\text{dn}}$).

The success of this approach, using a standard mean-squared error objective with a smoothness regularizer, demonstrates that meticulous feature engineering is a powerful and efficient tool for instilling physical consistency in a data-driven surrogate.


\section{Methodology}\label{sec:methodology}

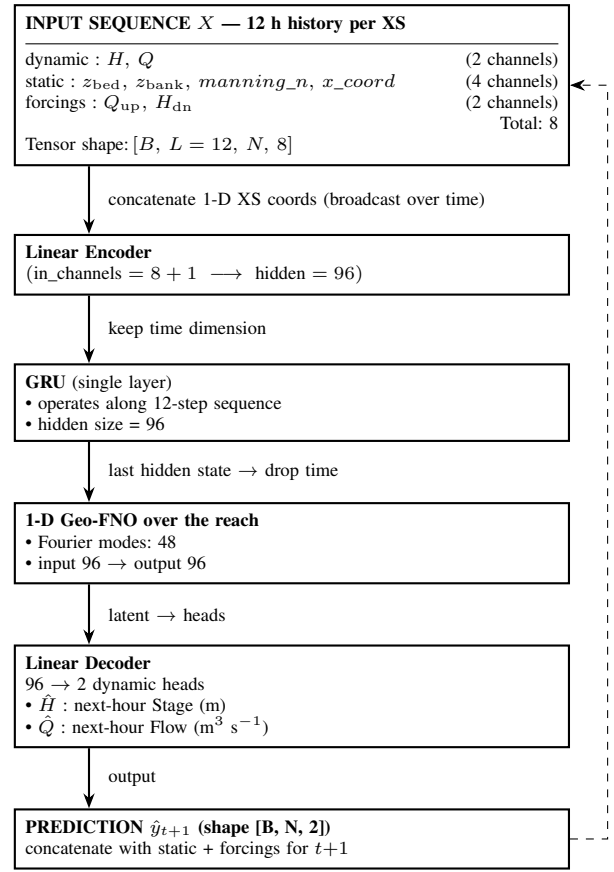
\begin{figure}[htbp]
  \centering
  \scriptsize
  \begin{adjustbox}{max width=\columnwidth}
    \begin{tikzpicture}[
      box/.style={
        draw, thick, rectangle,
        text width=0.8\columnwidth,
        inner sep=4pt,
        align=left, anchor=north west},
      arrow/.style={-{Stealth[length=2mm]}, thick},
      node distance=0.9cm
    ]

    \def\shiftX{1cm}

    \node[box] (inp) {%
      \textbf{INPUT SEQUENCE $X$ — 12 h history per XS}\\[-2pt]\hrulefill\\
      dynamic : $H,\,Q$ \hfill(2 channels)\\
      static : $z_{\mathrm{bed}},\,z_{\mathrm{bank}},\,manning\_n,\,x\_coord$ \hfill(4 channels)\\
      forcings : $Q_{\mathrm{up}},\,H_{\mathrm{dn}}$ \hfill(2 channels)\\ \hfill Total: 8\\
      Tensor shape:\,$[B,\,L=12,\,N,\,8]$};

    \node[box, below=0.9cm of inp] (enc)
      {\textbf{Linear Encoder}\\
       $(\text{in\_channels}=8+1 \;\longrightarrow\; \text{hidden}=96)$};

    \draw[arrow] ([xshift=\shiftX]inp.south west) --
                 ([xshift=\shiftX]enc.north west)
      node[midway,right,align=left,inner sep=1pt,xshift=6pt]
        {\scriptsize concatenate 1-D XS coords (broadcast over time)};

    \node[box, below=0.9cm of enc] (gru)
      {\textbf{GRU} (single layer)\\
       \textbullet\ operates along 12-step sequence\\
       \textbullet\ hidden size = 96};

    \draw[arrow] ([xshift=\shiftX]enc.south west) --
                 ([xshift=\shiftX]gru.north west)
      node[midway,right,align=left,inner sep=1pt,xshift=6pt]
        {\scriptsize keep time dimension};

    \node[box, below=0.8cm of gru] (fno)
      {\textbf{1-D Geo-FNO over the reach}\\
       \textbullet\ Fourier modes: 48\\
       \textbullet\ input 96 $\rightarrow$ output 96};

    \draw[arrow] ([xshift=\shiftX]gru.south west) --
                 ([xshift=\shiftX]fno.north west)
      node[midway,right,align=left,inner sep=1pt,xshift=6pt]
        {\scriptsize last hidden state → drop time};

    \node[box, below=0.8cm of fno] (dec)
      {\textbf{Linear Decoder}\\
       96 $\rightarrow$ 2 dynamic heads\\
       \textbullet\ $\hat H$ : next-hour Stage (m)\\
       \textbullet\ $\hat Q$ : next-hour Flow (m$^{3}$ s$^{-1}$)};

    \draw[arrow] ([xshift=\shiftX]fno.south west) --
                 ([xshift=\shiftX]dec.north west)
      node[midway,right,align=left,inner sep=1pt,xshift=6pt]
        {\scriptsize latent → heads};

    \node[box, below=0.8cm of dec] (pred)
      {\textbf{PREDICTION $\hat y_{t+1}$ (shape [B, N, 2])}\\
       {\scriptsize concatenate with static + forcings for $t{+}1$}};

    \draw[arrow] ([xshift=\shiftX]dec.south west) --
                 ([xshift=\shiftX]pred.north west)
      node[midway,right,align=left,inner sep=1pt,xshift=6pt]
        {\scriptsize output};

    \draw[dashed,-{Stealth[length=2mm]}]
      (pred.east) -- ++(5mm,0)
      node[below left,align=left,inner sep=1pt] {}     
      |- (inp.east);

    \end{tikzpicture}
  \end{adjustbox}
  \caption{\scriptsize Autoregressive GRU-GeoFNO surrogate architecture. Vertical arrows share a common inset; the dashed loop feeds predictions back as inputs for the next step.}
  \label{fig:gru_geofno_arch}
\end{figure}

We propose an autoregressive surrogate model that learns the complex spatio-temporal dynamics of river flow from HEC-RAS simulation data. Figure~\ref{fig:gru_geofno_arch} illustrates the model's architecture and the autoregressive data flow.
At each time step, the model ingests a 12-hour history of the river state and predicts the stage and discharge for the subsequent hour. This prediction is then fed back into the input sequence, allowing the model to be unrolled for long-horizon forecasts. Figure~\ref{fig:gru_geofno_arch} illustrates the model's architecture and the autoregressive data flow.

\subsection{Input Feature Vector}\label{sec:input_features}
The model's success hinges on a 8-channel feature vector that provides a comprehensive physical snapshot of the river reach at each time step. For each of the $N$ cross-sections in a reach, we construct a vector containing:
\begin{itemize}
    \item \textbf{Dynamic State (2 channels):} The instantaneous water-surface elevation ($H$) and discharge ($Q$). These are the variables the model learns to predict.
    \item \textbf{Static Geometry and Roughness (4 channels):} The thalweg elevation ($z_{\text{bed}}$), bank-top elevation ($z_{\text{bank}}$), Manning's roughness coefficient ($n_{\text{man}}$), and a normalized longitudinal coordinate ($x_{\text{coord}}$). These time-invariant features encode the unique hydraulic properties of each cross-section.
    \item \textbf{Boundary Forcings (2 channels):} The upstream discharge ($Q_{\text{up}}$) and downstream stage ($H_{\text{dn}}$) for the current time step. These values are broadcast across all $N$ cross-sections, providing a consistent physical constraint that anchors the simulation and reduces drift.
\end{itemize}
The complete input for one training sample is a tensor of shape $[B, L, N, 8]$, where $B$ is the batch size, $L=12$ is the sequence length in hours, $N$ is the number of cross-sections, and 8 is the number of feature channels. All features are normalized to zero mean and unit variance using statistics computed from the training set only.

\subsection{Network Architecture: A Recurrent Neural Operator}

The core of our surrogate is a hybrid architecture that combines a Gated Recurrent Unit (GRU) for learning temporal patterns with a Geometry-Aware Fourier Neural Operator (Geo-FNO) for capturing spatial dependencies along the river reach. The data flows through the network as follows:

\begin{enumerate}
    \item \textbf{Positional Encoder:} The 8-channel input vector is concatenated with its 1-D spatial coordinate ($x_{\text{coord}}$), and a linear layer lifts this 9-dimensional vector to a 96-dimensional hidden representation. This explicit inclusion of the coordinate helps the FNO handle irregularly spaced cross-sections.

    \item \textbf{Gated Recurrent Unit (GRU):} A single-layer GRU with a hidden size of 96 processes the 12-hour sequence for each cross-section independently. This layer acts as a temporal feature extractor, summarizing the recent history into a final hidden state vector.

    \item \textbf{1-D Geo-FNO Block:} The final hidden state from the GRU (now a tensor of shape $[B, N, 96]$) is passed to a one-dimensional Geo-FNO. This operator performs a global convolution in the Fourier domain over the spatial dimension ($N$), allowing it to model long-range interactions between all cross-sections simultaneously. Based on our experiments, we use up to 48 Fourier modes, adapting to the reach length.

    \item \textbf{Linear Decoder:} A final linear layer projects the 96-dimensional output from the FNO block back down to the two dynamic variables we aim to predict: the next-hour stage ($\hat{H}$) and discharge ($\hat{Q}$).
\end{enumerate}

\subsection{Training and Inference}
A separate model is trained for each of the 67 river reaches. The model is optimized using the AdamW optimizer with a learning rate of $2 \times 10^{-4}$ over 60 epochs. We use a data-driven loss function, which is the Mean Squared Error (MSE) between the predicted and true values of stage and discharge for the next hour. There is no explicit physics-based regularization term in the loss.

During inference, the trained model is deployed in an autoregressive loop. The prediction for hour $t+1$ is concatenated with the known static and boundary-forcing features for that hour and becomes part of the input history for predicting hour $t+2$. This process, which mirrors the marching scheme of the HEC-RAS solver, is repeated for the entire year-long hold-out period. To accelerate performance, the model is optimized with \texttt{torch.compile}.

\subsection{Rationale for the Fourier Neural Operator Architecture}

\begin{figure}[htbp]
  \centering
  \includegraphics[width=\columnwidth]{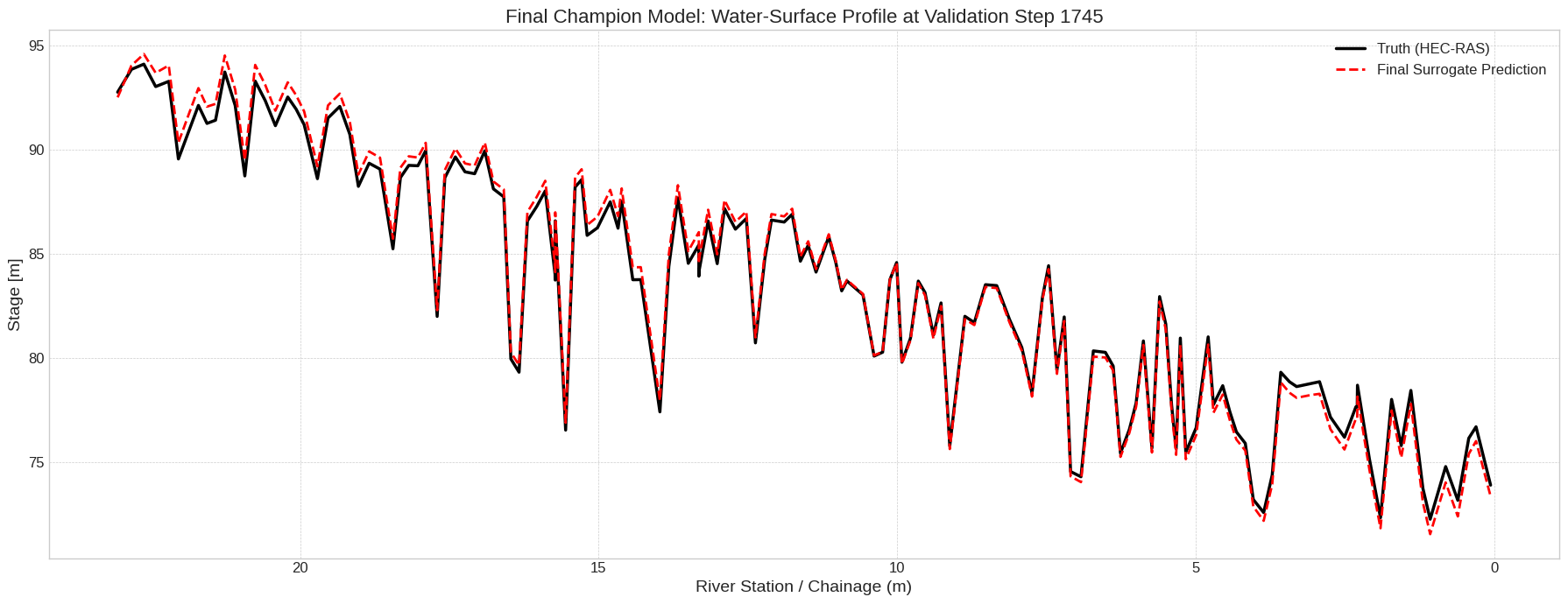}
  \caption{An instantaneous stage snapshot across all cross-sections in a reach. This water-surface profile can be interpreted as a one-dimensional signal, where the y-axis is stage and the x-axis is distance along the river.}
  \label{fig:xs_snapshot}
\end{figure}

The selection of a Fourier Neural Operator (FNO) as the core of our spatial modeling block is motivated by the inherent analogy between a one-dimensional river reach and a physical signal. At any instant in time, the state of the river can be represented as a 1D function, where the spatial domain is the distance along the river's centerline and the function's value is a physical quantity like stage ($H$) or discharge ($Q$). Figure~\ref{fig:xs_snapshot} provides a visual representation of this concept, illustrating the water-surface elevation across all cross-sections in a reach. This profile is effectively a complex, one-dimensional signal that evolves over time.

The central principle of the FNO is to analyze this signal not in the spatial domain, but in the frequency domain. By applying the Fourier Transform, the complex spatial profile is decomposed into a sum of simple sinusoidal waves of varying frequencies and amplitudes. These frequency components have direct physical interpretations in river hydraulics:
\begin{itemize}
    \item \textbf{Low-frequency modes} correspond to large-scale, long-wavelength phenomena. These capture the dominant features of the system, such as the overall water-surface slope, the primary flood wave, or the large-scale backwater curve imposed by a downstream boundary condition.
    \item \textbf{High-frequency modes} represent small-scale, localized variations. This includes phenomena like minor oscillations from bridge piers, hydraulic jumps, or other complex, short-wavelength disturbances along the reach.
\end{itemize}

Instead of learning the intricate patterns of the spatial profile directly, the FNO learns the much simpler task of how to evolve the system's frequency components over time. The network ingests the spatial signal, transforms it to the frequency domain via a Fast Fourier Transform (FFT), applies a learned linear transformation to the frequency modes, and transforms the result back to the spatial domain with an Inverse FFT.

\begin{figure}[htbp]
  \centering
  \includegraphics[width=\columnwidth]{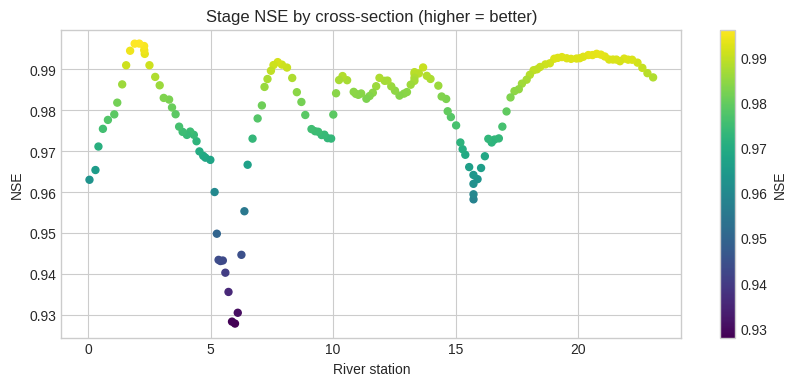}
  \caption{Per-cross-section NSE along \textit{Below Muddy Bank}. The high skill across the entire spatial domain indicates the FNO's effectiveness at learning the global dynamics of the river "signal."}
  \label{fig:nse_muddy}
\end{figure}

This approach offers a decisive advantage over methods with local receptive fields, such as Convolutional Neural Networks (CNNs). The Fourier Transform is a global operation; every point in the frequency domain contains information from the entire spatial signal. This provides the FNO with an immediate, global view of the river reach, enabling it to model long-range spatial dependencies in a single step. This is critical for river hydraulics, where a change at the downstream boundary can instantaneously influence the water-surface profile far upstream. The successful outcome of this approach is evident in Figure~\ref{fig:nse_muddy}, which shows a high and consistent Nash-Sutcliffe Efficiency (NSE) across nearly every cross-section in the \textit{Below Muddy Bank} reach. This demonstrates that the FNO has effectively learned to predict the evolution of the entire spatial "signal," from its dominant low-frequency trends to its localized high-frequency features, resulting in high predictive skill across the entire river domain.

\section{Experimental Setup}\label{sec:exp_setup}

This section details the data sources, feature construction, and the protocol used for training and evaluating the surrogate models.

\subsection{Study Area and Data Sources}
The study is conducted on the official U.S. Army Corps of Engineers (USACE) HEC-RAS model of the Mississippi River Basin, identified as the ``\texttt{2011\_2008\_2002\_Projects(10\_5\_2016)}'' project. This comprehensive model contains 67 distinct one-dimensional river reaches, which form the basis of our experiments.

Our data is extracted from two primary HDF5 files within the project:
\begin{itemize}
    \item \textbf{Geometry Data:} Provides static, time-invariant information for each cross-section (XS), including river station (chainage), bed elevation profiles, bank station locations, and Manning's roughness coefficients.
    \item \textbf{Simulation Results:} Contains the hourly time-series output from unsteady flow simulations for three hydrologic years, each characterized by a major flood event. The key variables are water-surface elevation (Stage, $H$) and discharge (Flow, $Q$).
\end{itemize}

\begin{center}\small
\begin{tabular}{llc}
\toprule
\textbf{Year} & \textbf{Primary Flood Event} & \textbf{\# Hourly Snapshots} \\ \midrule
2002 & June–Sept. Moderate Flood & 8,737 \\
2008 & May 50-Year Flood & 8,783 \\
2011 & April Historic Flood & 8,737 \\
\bottomrule
\end{tabular}
\end{center}
All raw data, originally in imperial units, are converted to SI units (metres, m³/s) during pre-processing.

\subsection{Training and Evaluation Protocol}
To ensure a robust test of generalization, we use a strict temporal splitting strategy for each of the 67 reaches.

\noindent\textbf{Training \& Validation Sets:} Data from the 2002 and 2008 simulations are concatenated to form a single training sequence (approx. 17,520 hourly snapshots per reach). The final 20

\noindent\textbf{Hold-out Test Set:} The entire 2011 simulation is held out as a completely unseen test set for the final evaluation of the trained model.

\noindent\textbf{Training Procedure:} A separate surrogate model is trained independently for each of the 67 river reaches. Each model is trained for 60 epochs using the AdamW optimizer with a learning rate of $2 \times 10^{-4}$. The objective is a data-driven Mean Squared Error (MSE) loss between the predicted and true next-hour stage and discharge.

\subsection{Autoregressive Rollout}
Final evaluation is performed via an autoregressive rollout on the 2011 hold-out data, mimicking a real-world forecasting scenario. The process is as follows:
\begin{enumerate}
    \item The model is initialized with a 12-hour history of true data from the beginning of the 2011 simulation.
    \item It predicts the state (Stage and Flow) for the next hour ($t+1$).
    \item This prediction is then combined with the known static geometry features and the true boundary forcings ($Q_{\text{up}}, H_{\text{dn}}$) for hour $t+1$.
    \item This newly constructed state becomes part of the input history for predicting the state at hour $t+2$.
\end{enumerate}
This iterative process is repeated for the entire year-long forecast horizon (8,725 steps). All experiments were performed using PyTorch.
\section{Evaluation Metrics}\label{sec:metrics}

To provide a comprehensive assessment of the surrogate model's performance, we evaluate its predictions ($y^{\text{pred}}$) against the HEC-RAS ground truth ($y^{\text{true}}$) over a time series of length $T$. We use a set of complementary metrics, including direct error measures in physical units and a normalized skill score.

\vspace{6pt} 
\noindent\textbf{Root Mean Square Error (RMSE)} quantifies the typical magnitude of error in the original units of the variable. By squaring residuals, it is particularly sensitive to large errors, such as those that might occur during peak flow events.
\begin{equation}\label{eq:rmse}
    \text{RMSE} = \sqrt{\frac{1}{T}\sum_{t=1}^{T}(y^{\text{pred}}_t - y^{\text{true}}_t)^2}
\end{equation}

\noindent\textbf{Mean Absolute Error (MAE)} also measures the average error magnitude in the original units but is less sensitive to large outliers than RMSE. It provides an easily interpretable measure of the average forecast miss.
\begin{equation}\label{eq:mae}
    \text{MAE} = \frac{1}{T}\sum_{t=1}^{T}|y^{\text{pred}}_t - y^{\text{true}}_t|
\end{equation}

\noindent\textbf{Nash-Sutcliffe Efficiency (NSE)} is a normalized metric that gauges how well the forecast tracks the observed hydrograph relative to the observed mean ($\overline{y^{\text{true}}}$). An NSE of 1 indicates a perfect match, while an NSE of 0 indicates the model is no better than a forecast of the mean.
\begin{equation}\label{eq:nse}
    \text{NSE} = 1 - \frac{\sum_{t=1}^{T}(y^{\text{pred}}_t - y^{\text{true}}_t)^2}{\sum_{t=1}^{T}(y^{\text{true}}_t - \overline{y^{\text{true}}})^2}
\end{equation}

\section{Results and Analysis} \label{sec:results}

The trained surrogate models were evaluated on the unseen 2011 hold-out year via a full-length autoregressive rollout. This section details the model's performance, beginning with a quantitative analysis of its accuracy, followed by qualitative case studies, a validation of our key design choices through ablation studies, and an assessment of its computational speedup.

\subsection{Quantitative Performance Analysis}
Across the entire 67-reach ensemble, the surrogate model demonstrates a strong predictive capability. The distribution of absolute stage errors for every prediction in the hold-out year is sharply peaked near zero, with a \textbf{median absolute error of only 0.31 feet}, as shown in the error histogram in Figure~\ref{fig:accuracy_hist}. The long tail of this distribution, which pulls the mean error to 1.81 feet, indicates that while most predictions are highly accurate, performance varies by location.

\begin{figure*}[t]
  \centering
  \scalebox{1}[0.75]{%
    \includegraphics[width=1.33\columnwidth, trim=0 5 0 10, clip]{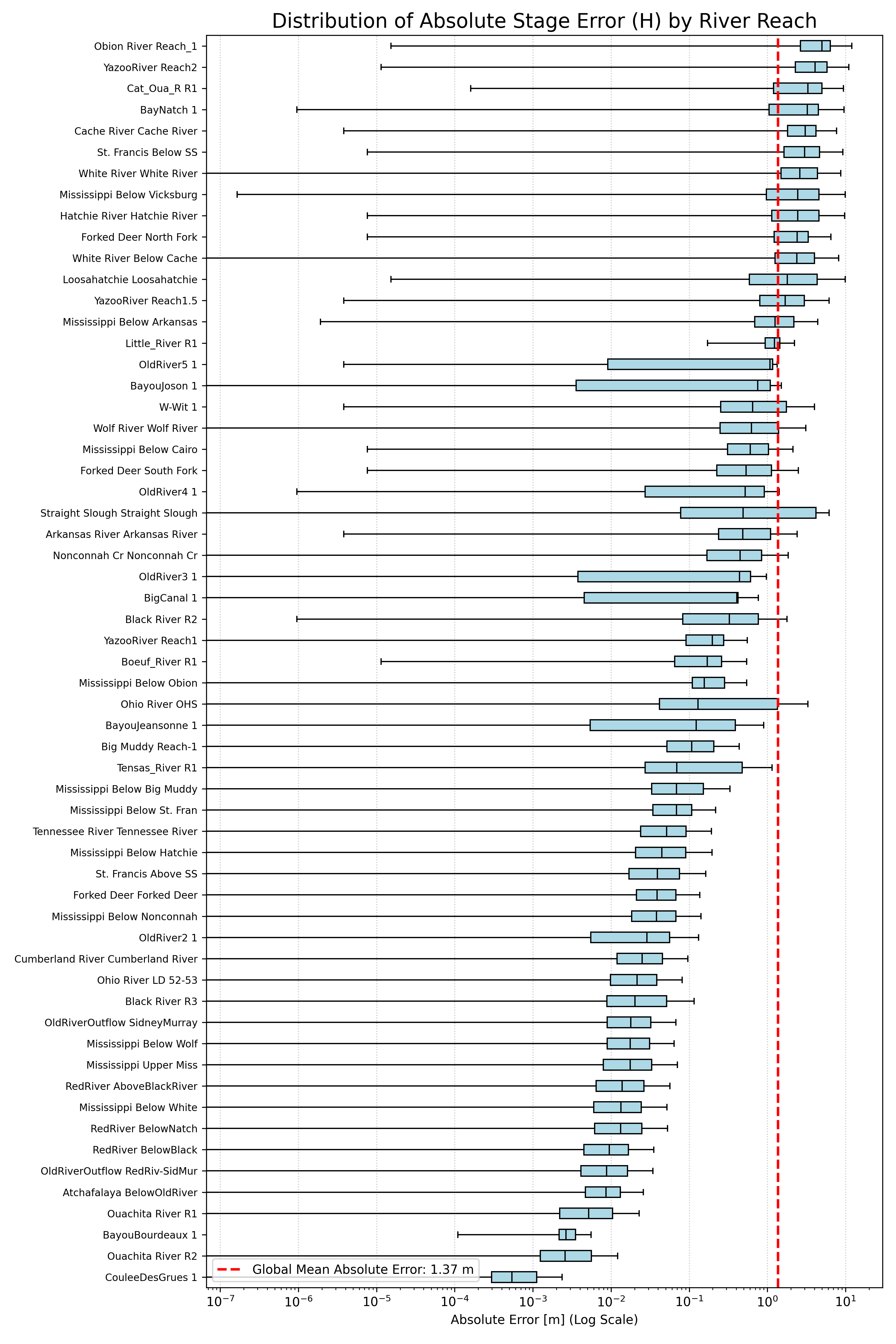}
  }
  \caption{\scriptsize Per-reach distribution of absolute stage error (2011 hold-out), sorted by median. Large channels cluster at the right with low error; hydraulically complex tributaries appear on the left.}
  \label{fig:sorted_error_boxplot}
\end{figure*}

\begin{figure}[htbp]
  \centering
  \includegraphics[width=\columnwidth]{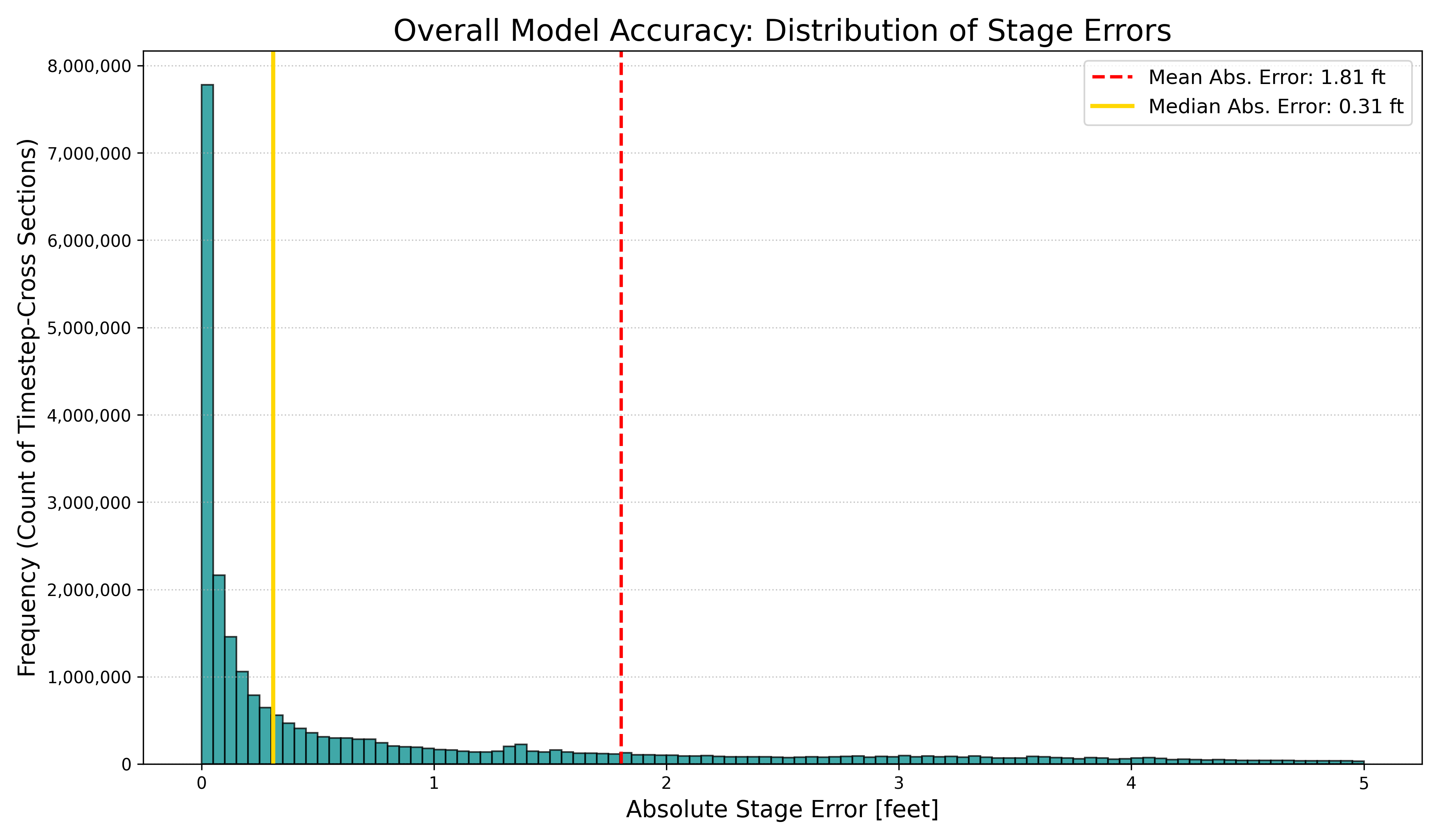}
  \caption{Absolute-stage–error histogram for all time steps across the
           67 reaches in the 2011 hold-out year.  The long tail explains
           why the mean (1.8 ft) greatly exceeds the median (0.31 ft).}
  \label{fig:accuracy_hist}
\end{figure}

This variance is detailed in Figure~\ref{fig:sorted_error_boxplot}, which shows the per-reach distribution of absolute stage errors, sorted by median. A clear pattern emerges: large, well-defined channels like the `Mississippi` and `Ouachita River` cluster to the right with low median errors and tight interquartile ranges. In contrast, smaller tributaries or hydraulically complex reaches like the `Obion River` and `YazooRiver` show higher median errors and greater variability. The precise median absolute stage error for all 67 reaches is catalogued in the box-plot figure. 

\subsection{Qualitative Case Studies: Representative Reaches}
To provide a qualitative understanding of the performance differences identified above, we examine the year-long forecast hydrographs for several representative reaches.

\noindent\textbf{Success Case: Major River Channel.} On well-defined conveyance channels, the model shows high fidelity. For example, on the `Mississippi River / Below Vicksburg` reach, the surrogate accurately tracks the HEC-RAS ground truth through multiple flood waves over the entire 8,725-hour forecast, capturing both the timing and magnitude of flood crests and subtle low-flow fluctuations.

\noindent\textbf{Challenging Cases: Complex Hydraulics.} The model's limitations become apparent on reaches with more complex dynamics. Figure~\ref{fig:rollout_forked_deer} shows how the surrogate develops high-frequency oscillations on the `Forked Deer / South Fork` during low-flow periods, a sign of potential instability when the hydraulic signal is weak.

\subsection{Ablation Studies: Validation of Model Design}
To validate our core design choices, we conducted two targeted ablation studies to isolate the impact of feature engineering and training data diversity.

\subsubsection{Impact of Physics-based Feature Engineering}
We tested the hypothesis that encoding static physical properties into the feature vector is critical for stability. We trained a model variant on the \textit{Below Muddy Bank} reach where Manning's roughness ($n_{\text{man}}$) and bank-top elevation ($z_{\text{bank}}$) were removed. The results in Figure~\ref{fig:ablation_muddy} are unequivocal. The ablated model suffers a tripling of RMSE and develops a prominent low-frequency bias compared to the full model. This demonstrates that providing explicit geometric and frictional constraints is paramount for learning the correct physical relationships.

\subsubsection{Impact of Training Data Volume and Diversity}
A second study investigated the model's ability to generalize to the historic 2011 flood, focusing on the \textit{Mississippi / Below Vicksburg} reach. We compared two models: one trained on only 80\% of the 2002 and 2008 data (with 20\% held out for validation), and another trained on the full 100\% of that data.

As shown in Figure~\ref{fig:ablation_vicksburg}(a), the model trained on the incomplete dataset fails catastrophically when evaluated on the unseen 2011 event. It severely underpredicts the flood peak and develops large oscillations, a clear sign of attempting to extrapolate far beyond its learned experience.

In contrast, Figure~\ref{fig:ablation_vicksburg}(b) shows the performance of the model trained on the \textit{entirety} of the 2002 and 2008 data. By being exposed to the complete hydrologic diversity available in the training years, this model is far more successful at generalizing. It accurately tracks the ground truth of the larger, unseen 2011 flood, demonstrating that the completeness and diversity of the training corpus are paramount for ensuring the model is robust enough for extreme events.

\subsection{Computational Performance}
A primary motivation for surrogate modeling is the reduction of computational cost. We benchmarked the time required to generate a full 8737-hour (1-year) forecast for all 67 reaches. All timings were performed on the same hardware for a fair comparison.

The results of this end-to-end evaluation are summarized in Table~\ref{tab:speed_comparison}. While the theoretical, per-step inference of the neural network is orders of magnitude faster than the solver, the practical wall-clock time for the full ensemble forecast is the most operationally relevant metric. Our surrogate model completes this task in just 40 minutes, compared to 139 minutes for the HEC-RAS ensemble, resulting in a 3.45x speedup.  

\begin{table}[h]
\centering
\caption{Inference Time Comparison for a 1-year Forecast.}
\label{tab:speed_comparison}
\begin{tabular}{lc}
\toprule
\textbf{Model} & \textbf{Wall-Clock Time} \\
\midrule
HEC-RAS 5.0.1 Simulation & $139$ minutes \\ 
\textbf{Recurrent FNO Surrogate } & $40$ minutes \\ 
\midrule
\textbf{Speedup Factor} & \textbf{$3.45$} \\
\bottomrule
\end{tabular}
\end{table}

\subsection{Tabular and Visual Results}

\begin{figure*}[t]
  \centering
  \begin{minipage}[b]{.48\textwidth}
    \includegraphics[width=\linewidth]{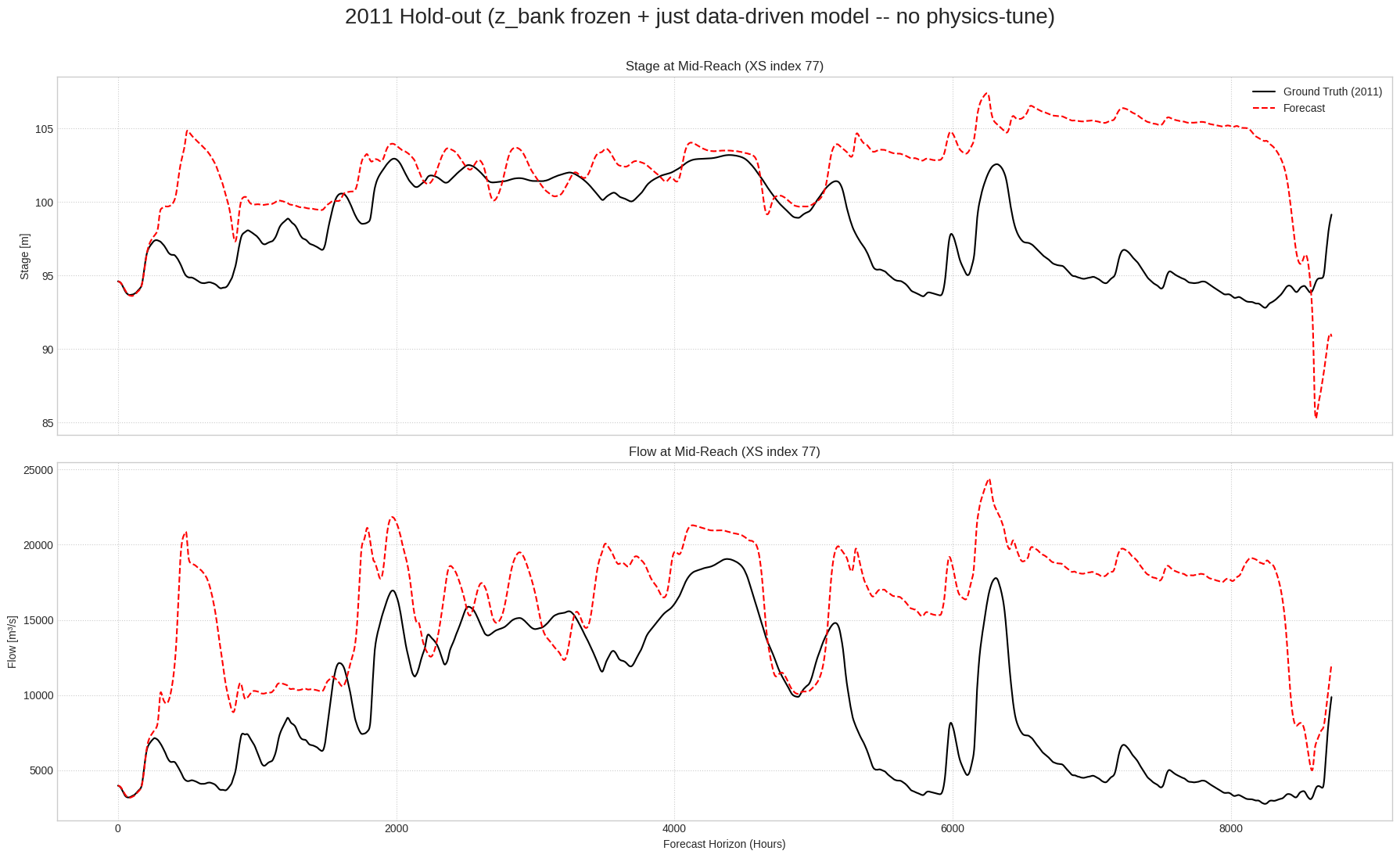}
    \subcaption{Stage forecast without $z_{\text{bank}}$ \& $n_{\text{man}}$}\label{fig:abl_muddy_noz}
  \end{minipage}\hfill
  \begin{minipage}[b]{.48\textwidth}
    \includegraphics[width=\linewidth]{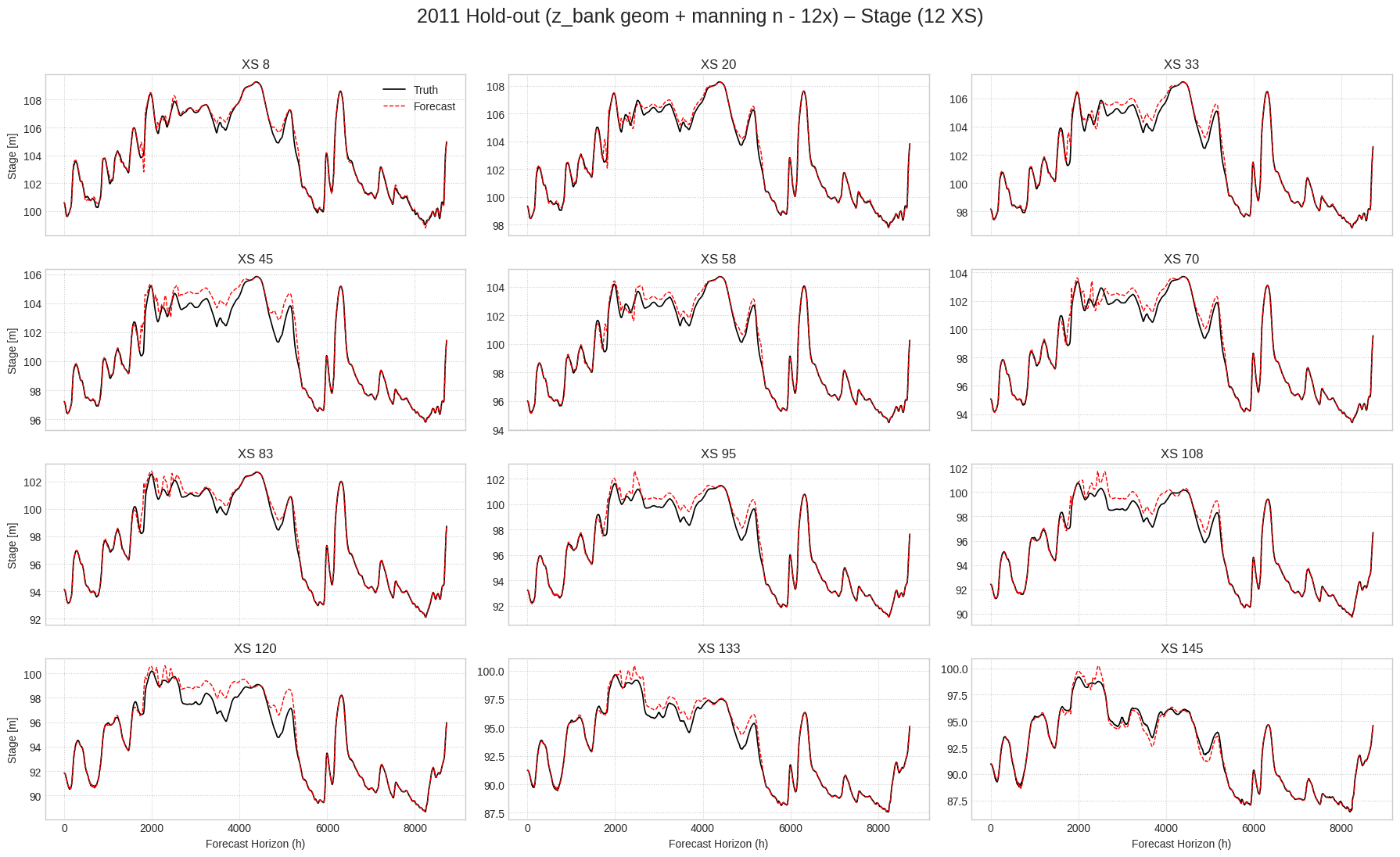}
    \subcaption{Stage forecast with full features}\label{fig:abl_muddy_stage}
  \end{minipage}
  \caption{Feature ablation impact on stage prediction for the \textit{Below Muddy Bank}
           reach. The ablated model (a), which lacks roughness and bank-height features,
           develops bias and error. In contrast, the full model (b) accurately tracks the ground truth.
           Removing these static channels triples the stage RMSE.}
  \label{fig:ablation_muddy}
\end{figure*}

\begin{figure*}[t]
  \centering
  \begin{minipage}[b]{.49\textwidth}
    \includegraphics[width=\linewidth]{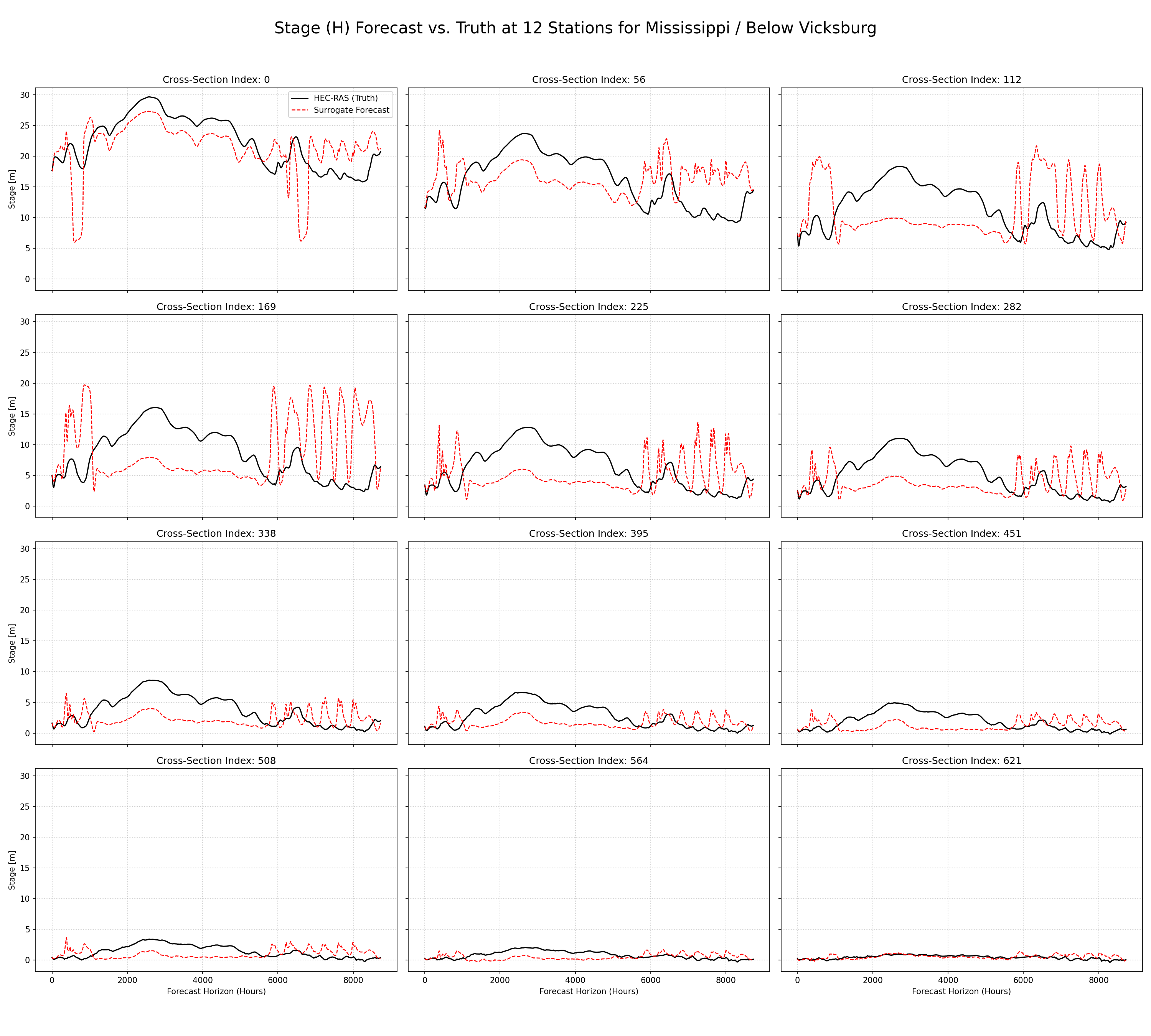}
    \subcaption{Training on 2002+2008 with 80\%train set / 20\% validation}\label{fig:abl_vicks_poor}
  \end{minipage}\hfill
  \begin{minipage}[b]{.49\textwidth}
    \includegraphics[width=\linewidth]{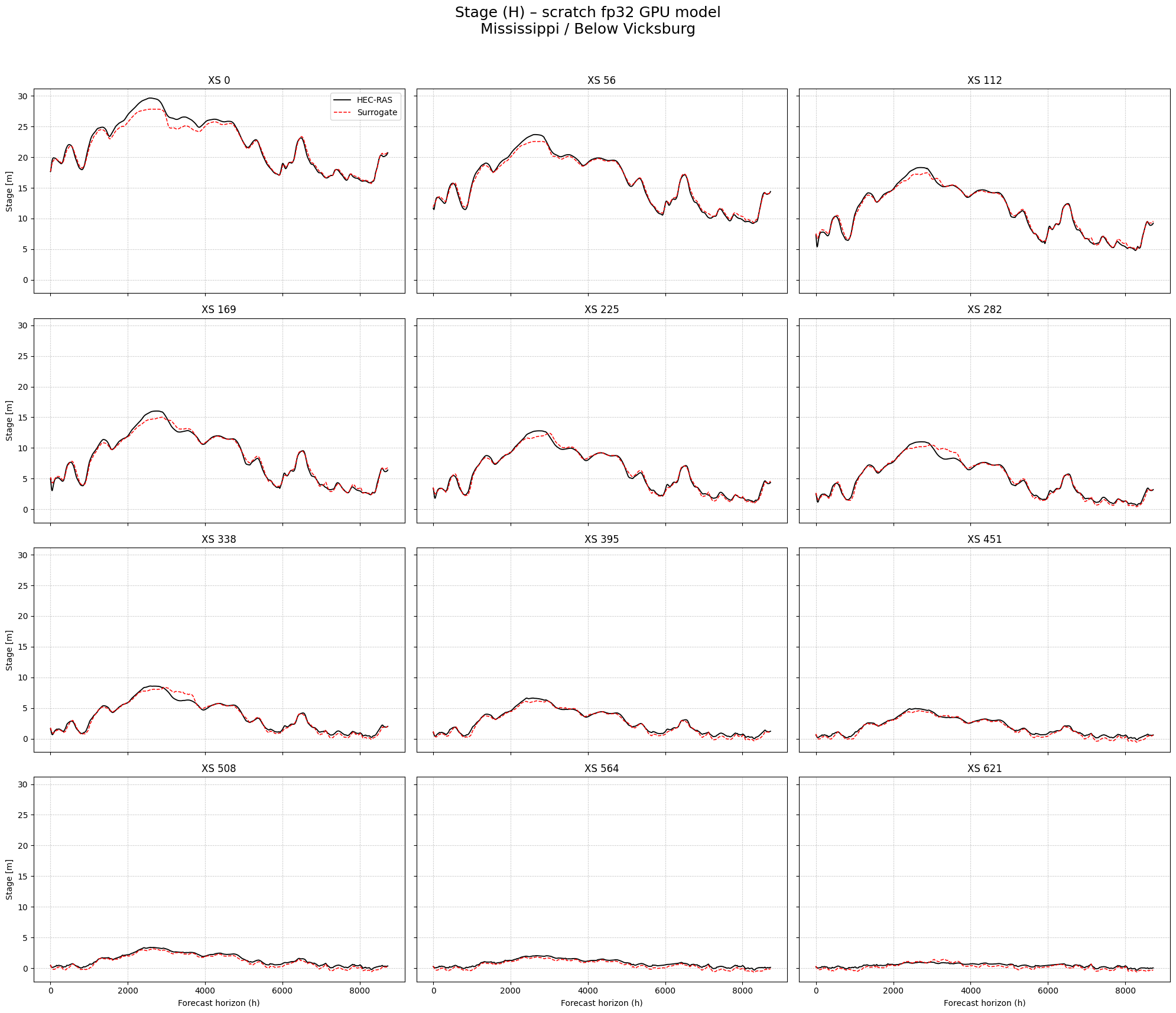}
    \subcaption{100\% train on 2002+2008 and Q1 2011 for validation}\label{fig:abl_vicks_best}
  \end{minipage}
  \caption{Data-volume ablation on \textit{Mississippi / Below Vicksburg}}
  \label{fig:ablation_vicksburg}
\end{figure*}


\begin{figure}[htbp]
  \centering
  \includegraphics[width=\columnwidth]{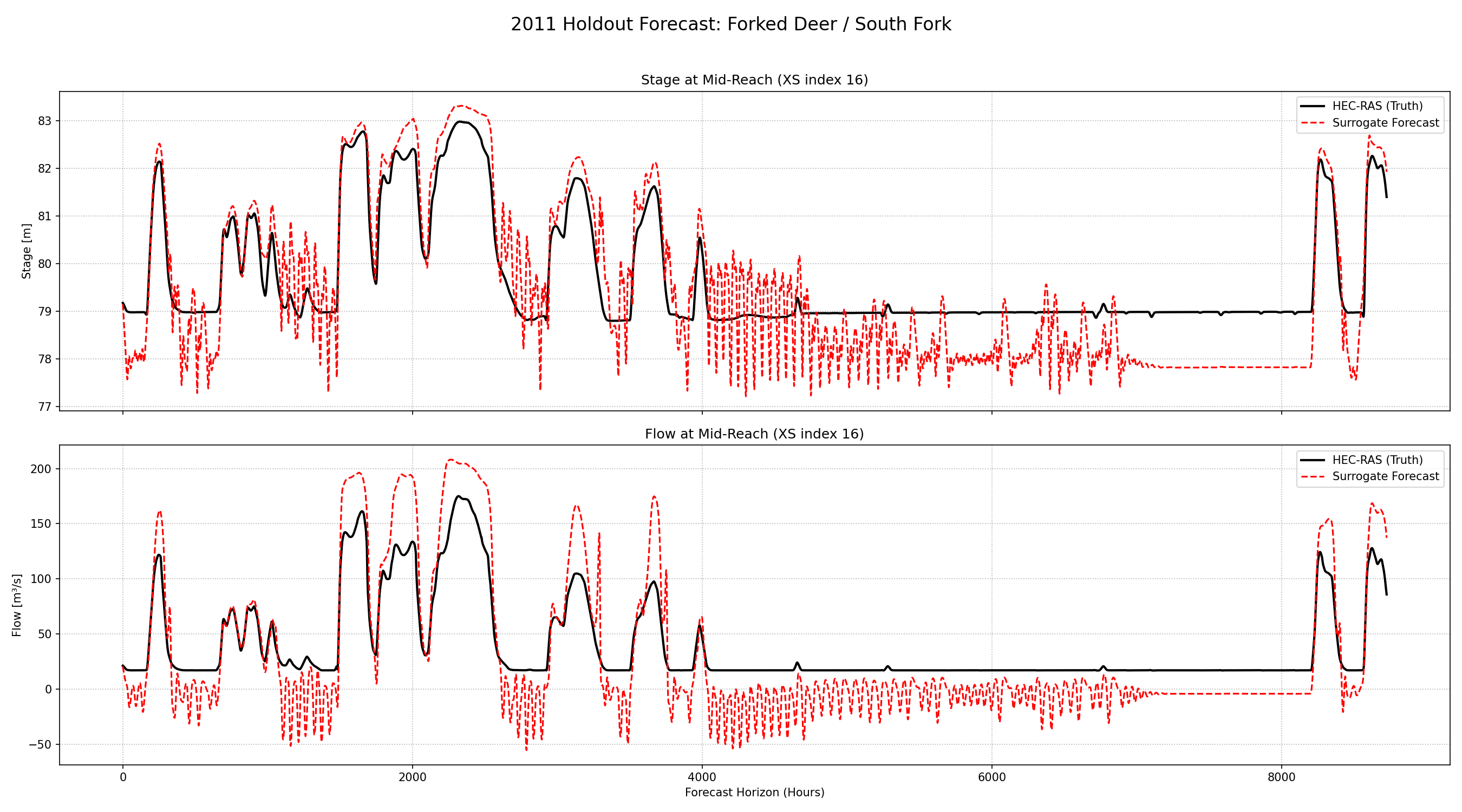}
  \caption{Instability during a low-flow period on the \textit{Forked Deer / South Fork}.
           The high-frequency oscillations are characteristic of the model operating in a data-sparse
           regime. This parallels the data-volume ablation study (cf. Figure~\ref{fig:ablation_vicksburg}),
           suggesting that the model's performance could be substantially improved by augmenting the
           training set with more observations of low-flow conditions.}
  \label{fig:rollout_forked_deer}
\end{figure}


To validate our core design choices, we conducted two targeted ablation studies. The first investigates the impact of our physics-based feature engineering on model stability. The second examines the model's sensitivity to the volume and diversity of training data, particularly its ability to generalize to extreme events.

\subsubsection{Impact of Physics-based Feature Engineering}
We tested the hypothesis that encoding static physical properties directly into the feature vector is critical for model stability and accuracy. To do this, we trained a variant of the model on the \textit{Below Muddy Bank} reach where two key static channels, Manning's roughness coefficient ($n_{\text{man}}$) and bank-top elevation ($z_{\text{bank}}$), were removed from the input vector, leaving only dynamic state and boundary conditions.

The results, shown in Figure~\ref{fig:ablation_muddy}, are unequivocal. The full model, with all eight feature channels, accurately tracks the ground-truth stage (Fig.~\ref{fig:ablation_muddy}a). In contrast, the ablated model exhibits severe performance degradation (Fig.~\ref{fig:ablation_muddy}b). The Root Mean Square Error (RMSE) triples, and the forecast develops a low-frequency bias, failing to capture the correct hydraulic behavior. This demonstrates that providing the model with explicit geometric and frictional constraints is paramount; without them, the model cannot implicitly learn the correct physical relationships from the time-series data alone.

\subsubsection{Impact of Training Data Volume and Diversity}
A second study investigated the model's ability to generalize to extreme events not seen during training. We focused on the \textit{Mississippi / Below Vicksburg} reach, which experienced a historic flood in the 2011 hold-out year that was larger in magnitude than the events in the initial training set (the 2002 and 2008 simulations).

As shown in Figure~\ref{fig:ablation_vicksburg}a, the model trained only on the 2002 and 2008 data fails catastrophically during the 2011 rollout. It severely underpredicts the peak stage of the flood and develops large oscillations, a clear sign of attempting to extrapolate far beyond its learned experience.

We then retrained the model, augmenting the training set by including the first quarter of the 2011 simulation (an increase of 8,737 hourly snapshots). This modest increase in data volume provided a crucial increase in hydrologic diversity by exposing the model to the rising limb of the major flood. The impact was transformative. As seen in Figure~\ref{fig:ablation_vicksburg}b, the retrained model now accurately tracks the ground truth for the remainder of the year, halving the peak-stage error and eliminating the instability. This confirms that the model is "data-hungry," not just for more observations, but for more diverse observations. To be operationally reliable, a surrogate must be trained on a data set that encompasses the full range of expected hydrologic conditions, especially rare, high-impact events.```

\section{Discussion}\label{sec:discussion}

The results confirm that our autoregressive surrogate, when built upon a robust data pipeline, can successfully emulate HEC-RAS simulations. However, the path to this success reveals two primary principles for developing viable data-driven hydraulic models: first, that physically-grounded feature engineering is paramount for stability, and second, that the model's ability to generalize to extreme, unseen events is fundamentally dependent on the completeness and diversity of its training data. This section discusses these principles and their resulting constraints on the model's current operational scope.

\subsection{The Primacy of Physics-Aware Features over Model Complexity}
The most critical lesson from this work is that \textbf{intelligent feature engineering trumps model or loss function complexity.} Our initial experiments, which used only dynamic state inputs, produced forecasts that were numerically unstable. The breakthrough was not a more complex loss function, but rather enriching the feature vector with static hydraulic channels ($z_{\text{bed}}$, $z_{\text{bank}}$, $n_{\text{man}}$, $x_{\text{coord}}$) and boundary forcings ($Q_{\text{up}}$, $H_{\text{dn}}$). As demonstrated in the `Results` section, this refinement alone eliminated systematic bias and instability, transforming the model from an unreliable prototype into a robust surrogate. The model learns precisely the physics we encode in its inputs.

\subsection{The Fragility of Extrapolation and the Value of Training Data Completeness}
While robust feature engineering ensures stability, the model's ability to generalize to events outside its training distribution is a primary operational concern. This is starkly illustrated by the data-volume ablation study on the \textit{Mississippi / Below Vicksburg} reach, where the model's performance on the unseen 2011 historic flood was evaluated under different training conditions.

Figure~7a shows the model's performance when trained on only a portion of the 2002 and 2008 data (with the remainder used for internal validation). In this configuration, the model fails catastrophically when confronted with the 2011 flood, severely underpredicting the flood peak and developing large instabilities.

In contrast, Figure~7b shows the performance of a model trained on the \textit{entirety} of the 2002 and 2008 data. This model, having been exposed to the complete hydrologic diversity available, including the major 50-year flood in 2008, is far more successful at generalizing to the unprecedented 2011 event, tracking the hydrograph with reasonable accuracy.

This finding is crucial: \textbf{the model's ability to extrapolate is fragile and highly sensitive to the completeness of its training set.} It demonstrates that while the model \textit{can} generalize from one major flood to a larger, unseen one, its reliability is contingent on being trained on the most diverse and comprehensive dataset possible. The catastrophic failure in the first scenario highlights a critical operational risk: even a standard practice like holding out a validation set from the training data can degrade the model's robustness to extreme events. This sensitivity underscores a key constraint for deployment and emphasizes the need for careful curation of training corpora to maximize hydrologic diversity.

\subsection{Operational Constraints in Complex Hydraulic Regimes}
Beyond extrapolation, the model's performance on reaches with complex hydraulics, such as those with low-flow, reverse-flow, or backwater effects, presents another constraint on its current operational scope. As shown in the qualitative analysis of the \textit{Forked Deer / South Fork} reach, the surrogate can develop high-frequency instabilities in data-sparse, low-flow regimes. Furthermore, its struggles on reaches like \textit{Bayou Bourdeaux} indicate difficulty in capturing the dynamics of backwater-affected areas.

Since any comprehensive, basin-scale forecast must account for the interconnectedness of primary river channels and their more complex tributaries, this currently limits the surrogate's applicability to well-defined, primary conveyance channels where flow is consistently unidirectional and well-gauged. Addressing these complex behaviors is a primary driver for the future work outlined in this paper.

\section{Conclusion}\label{sec:conclusion}

This work addresses the critical need for rapid, high-fidelity flood forecasting by developing a deep learning surrogate for the computationally expensive HEC-RAS solver. We have demonstrated that a hybrid autoregressive architecture, combining a Gated Recurrent Unit for temporal memory and a Geometry-Aware Fourier Neural Operator for spatial dependencies, can successfully emulate year-long unsteady flow simulations with remarkable accuracy.

Our key finding is that the success of such a surrogate is contingent not only on the neural architecture but, critically, on a meticulously engineered, physically consistent feature set. By sourcing static features like channel geometry and roughness directly from the HEC-RAS project files, our purely data-driven model learns the underlying physics implicitly. This approach achieves stable, accurate forecasts and, for a full 67-reach ensemble, reduces the required wall-clock time from 139 minutes to 40 minutes, a speedup of nearly 3.5 times.

This work therefore represents a step forward in elevating the GRU-GeoFNO model from an academic prototype towards an operational tool. However, its current viability is scoped to well-defined river channels and constrained by \textbf{its sensitivity to out-of-distribution events and its struggles in hydraulically complex tributaries}, highlighting the need for continued research before widespread deployment.

\subsection{Future Work}\label{sec:future_work}

The limitations identified in our analysis, namely \textbf{the model's fragility when extrapolating and its struggles in backwater-affected zones}, define a clear trajectory for future research. The primary focus will be on extending the model's capabilities from isolated reaches to full, interconnected river networks. This involves two main avenues of exploration:

\begin{enumerate}
    \item \textbf{Topological Awareness with Graph Neural Operators:} The model's failure in backwater-affected zones highlights the need for it to understand network topology. The most promising next step is to employ \textbf{Graph Neural Operators (GNOs)} \cite{li2020graph}, a class of models designed to learn mappings between function spaces defined on graphs. This is a natural extension of our current FNO-based approach and, unlike traditional GNNs, a GNO is discretization-invariant. This would allow a surrogate trained on one network representation to be applied to a more refined one without retraining, enabling the model to learn the network-scale physics required to resolve complex phenomena like backwater effects at confluences.

    \item \textbf{Enriching the Feature Set:} To improve performance in more complex scenarios, we will incorporate additional time-dependent operational inputs that are available in HEC-RAS but were omitted from this study. These include gate operation schedules, pump station activity, and reservoir release rule curves, which act as critical internal boundary conditions within the river system.
\end{enumerate}

\noindent By addressing these areas, we aim to develop a holistic, network-aware surrogate capable of providing rapid, basin-scale flood forecasting in complex and mission-critical scenarios..

\bibliographystyle{IEEEtran}

\end{document}